\documentclass{article} 
\usepackage{iclr2023_conference,times}
\usepackage{graphicx}

\usepackage{amsmath,amsfonts,bm}

\def\eqref#1{equation~\ref{#1}}

\def\1{\bm{1}}

\DeclareMathAlphabet{\mathsfit}{\encodingdefault}{\sfdefault}{m}{sl}
\SetMathAlphabet{\mathsfit}{bold}{\encodingdefault}{\sfdefault}{bx}{n}

\usepackage{hyperref}
\usepackage{url}
\newcommand{\latent}{\mathbf{z}_t}
\newcommand{\latenta}{\mathbf{h}_t}
\newcommand{\state}{\mathbf{s}_t}
\newcommand{\action}{\mathbf{a}_t}
\newcommand{\trans}{\mathbf{T}_t}
\newcommand{\latentnext}{\mathbf{z}_{t+1}}
\newcommand{\inflatentnext}{\tilde{\mathbf{z}}_{t+1}}
\usepackage{float}
\usepackage{algorithm}
\usepackage{algpseudocode}

\title{Learning parsimonious dynamics \\ for generalization in reinforcement learning}

\author{Tankred Saanum \\
Computational Principles of Intelligence Lab \\
Max-Planck-Institute for Biological Cybernetics \\
Tübingen, Germany \\
\And
Eric Schulz \\
Computational Principles of Intelligence Lab \\
Max-Planck-Institute for Biological Cybernetics \\
Tübingen, Germany \\
}

\iclrfinalcopy 
\begin{document}

\maketitle

\begin{abstract}

Humans are skillful navigators: We aptly maneuver through new places, realize when we are back at a location we have seen before, and can even conceive of shortcuts that go through parts of our environments we have never visited. Current methods in model-based reinforcement learning on the other hand struggle with generalizing about environment dynamics out of the training distribution. We argue that two principles can help bridge this gap: latent learning and parsimonious dynamics. Humans tend to think about environment dynamics in simple terms -- we reason about trajectories not in reference to what we expect to see along a path, but rather in an abstract latent space, containing information about the places' spatial coordinates. Moreover, we assume that moving around in novel parts of our environment works the same way as in parts we are familiar with. These two principles work together in tandem: it is in the latent space that the dynamics show parsimonious characteristics. We develop a model that learns such parsimonious dynamics. Using a variational objective, our model is trained to reconstruct experienced transitions in a latent space using locally linear transformations, while encouraged to invoke as few distinct transformations as possible. Using our framework, we demonstrate the utility of learning parsimonious latent dynamics models in a range of policy learning and planning tasks.

\end{abstract}

\section{Introduction}

Navigation comes easy to humans. We are able to maneuver through novel parts of our environments, self-locate by integrating over convoluted trajectories, and even come up with shortcuts that traverse areas of the environment we have never visited before. Two principles seem to drive these abilities: latent learning and parsimonious dynamics. Latent learning describes the ability to represent paths through our world not as we experience them literally, but in an abstract manner: That is, we reason about trajectories not in reference necessarily to what we expect to see along a path, but rather in an abstract latent space, containing information about the places' spatial coordinates \citep{tolman1948cognitive, constantinescu2016organizing}. These coordinates are themselves never experienced, but are useful representations constructed to reflect the structure of the environment we inhabit. Parsimonious dynamics describe the fact that the rules governing how state transitions work should be simple.  We assume that moving around in novel parts of our environment works the same way as in parts we are familiar with. These two principles work together in tandem: it is in the latent space that the dynamics show parsimonious characteristics.

We extend these ideas to the more general framework of learning latent dynamics models for reinforcement learning (RL). Recent advances in model-based RL have showcased the potential improvements in the performance and sample complexity that can be gained by learning accurate latent dynamics models \citep{deisenroth2011pilco, hafner2019dream, schrittwieser2020mastering}. These models summarize the transitions that the agent experiences by interacting with its environment in a low-dimensional latent code that is learnt alongside their dynamics. By learning such models, agents may for instance perform control by planning ahead in this low-dimensional state space \citep{hafner2019learning}, or, if the latent states contain useful information for policy learning, simply learn a policy over latent states rather than the original states \citep{ha2018recurrent, lee2020stochastic}. We employ the principle of parsimony to learn a latent dynamics model that is able to generalize about novel transitions, and whose latent states contain information about the topology of the environment. These latent state representations prove to be useful for policy learning, planning and future state prediction. Moreover, we show that one can learn such latent representations of the environment simply from encouraging the dynamics to be parsimonious, without supervision about the underlying latent structure.  


\begin{figure}[t!]
\begin{center}
\includegraphics[width=1\textwidth]{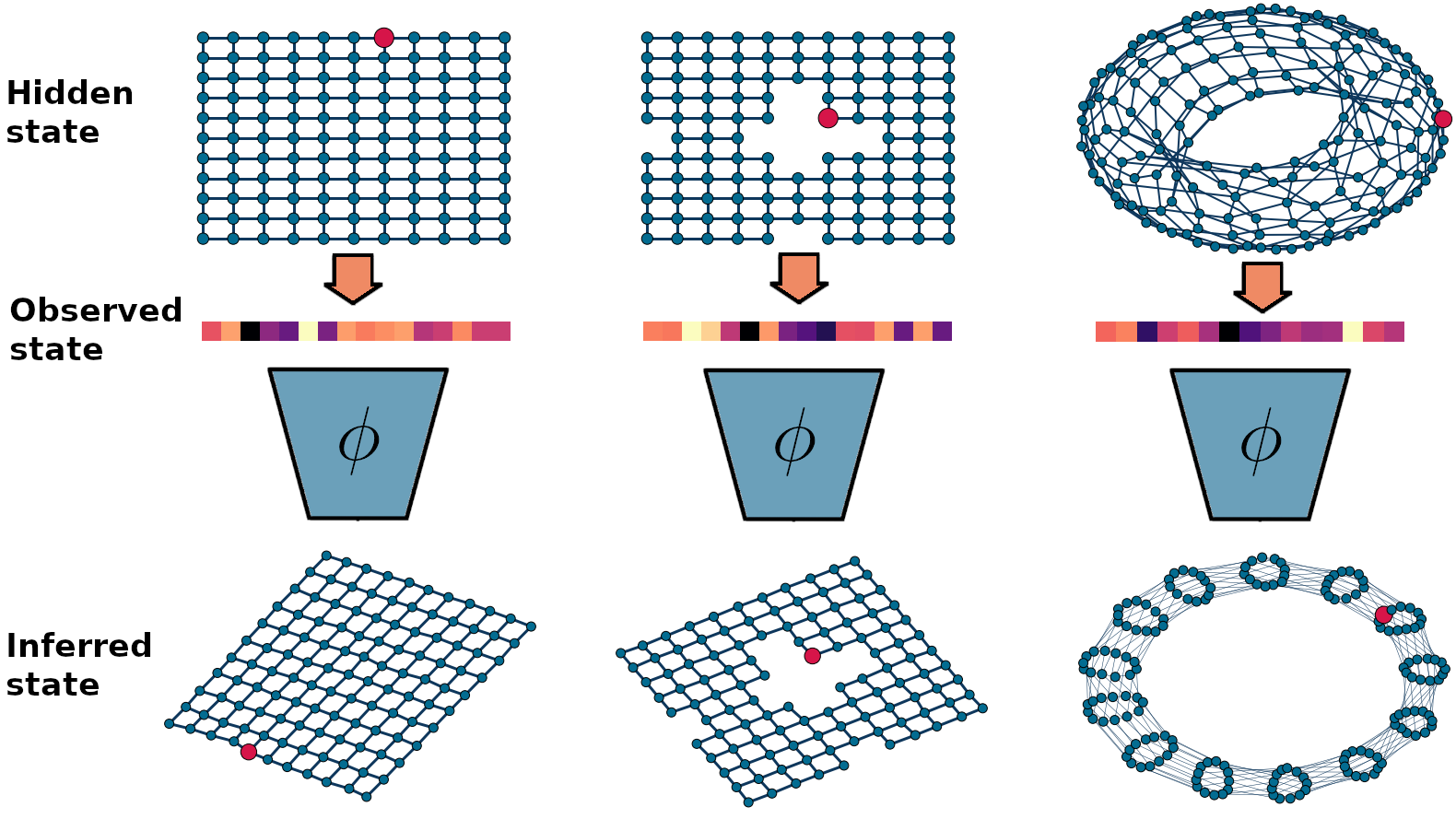}

\end{center}
\caption{Illustration of the learning problem. Environments can be embedded in an underlying low-dimensional space, and transitions can be described by a small set of linear transformations, like rotations and translations, information that is unavailable to the agent. The observation space $\mathcal{S}$ is not informative of the underlying topological structure of the environments, and must be learnt. Our model discovers a latent space $\tilde{\mathcal{Z}}$ where state-transitions can be described by a small number of learnt transformations.}

\end{figure}

We draw inspiration from group theory, the study of transformations that preserve symmetries \citep{kondor2008group}, to learn such latent spaces. Particularly, we adopt the framework of \cite{quessard} and \cite{caselles2019symmetry} where the interventions an agent can perform on its environment are treated as transformations belonging to a group, and transitioning between states through selecting actions is equivalent to transforming the source state with the action's corresponding group transformation. We extend this approach by summarizing a data set of experienced transitions invoking only a small set of different types of learned transformations. We hypothesize that a model that infers a small set of locally linear transformations to explain global transition dynamics should be able to generalize effectively about the transition dynamics of novel parts of the agent's environment, and that the latent state representations that result from embedding states onto such discovered manifolds are beneficial for policy learning. In the end, we show that our approach outperforms alternative dynamics and representation learning models in planning and policy learning tasks, as well as in an open-loop pixel prediction tasks based on the Deepmind Lab environment \citep{beattie2016deepmind}.



\section{Model}
\subsection{Preliminaries}

We assume that the environment is a Markov Decision Process (MDP) defined as the tuple $\langle \mathcal{S}, \mathcal{A}, \mathcal{R}, \mathcal{T}, \gamma\rangle$, where $\mathcal{S}$ is the state space, $\mathcal{A}$ is the set of actions, $\mathcal{T}$ is the transition function describing the probability of successor states given the current state-action tuple $\mathbf{s}_{t+1}\sim \mathcal{T}(\state, \action)$, $\mathcal{R}$ is the reward function and $\gamma$ is the discount factor. In every state, the agent selects an action according to the policy $\action \sim \pi(\state)$. The agent's goal is to learn a policy $\pi(\action\mid \state)$ or plan using knowledge of environment dynamics to maximize future discounted rewards $\mathop{\mathbb{E}}_{\mathcal{T}}\left[\sum_{t=0}^{\infty}\gamma^t\mathcal{R}(\state)\right]$.

The original state space of the MDP may be high-dimensional and difficult to perform RL on. We hypothesize that constructing a low-dimensional latent space that exhibits parsimonious dynamics is beneficial for RL in at least two ways: i) Learning a policy on the latent space $\pi(\mathbf{a}\mid \latent)$ should be easier since the latent states are organized such that they match the underlying, hidden topology of the environment. ii) Planning should be possible with less experience: the gains we seek to make here lie in exploiting the knowledge of the transformations that describe how the state variable changes with our actions. For instance, if the agent learns that all state-transitions can be described by a small set of transformations of the source state depending on the action, we can correctly generalize about what the next latent state will be simply if we can predict what the appropriate transformation of our current latent state is, given the action the agent selected. 

\subsection{Model components}


\begin{figure}[t!]
\begin{center}
\includegraphics[width=1\textwidth]{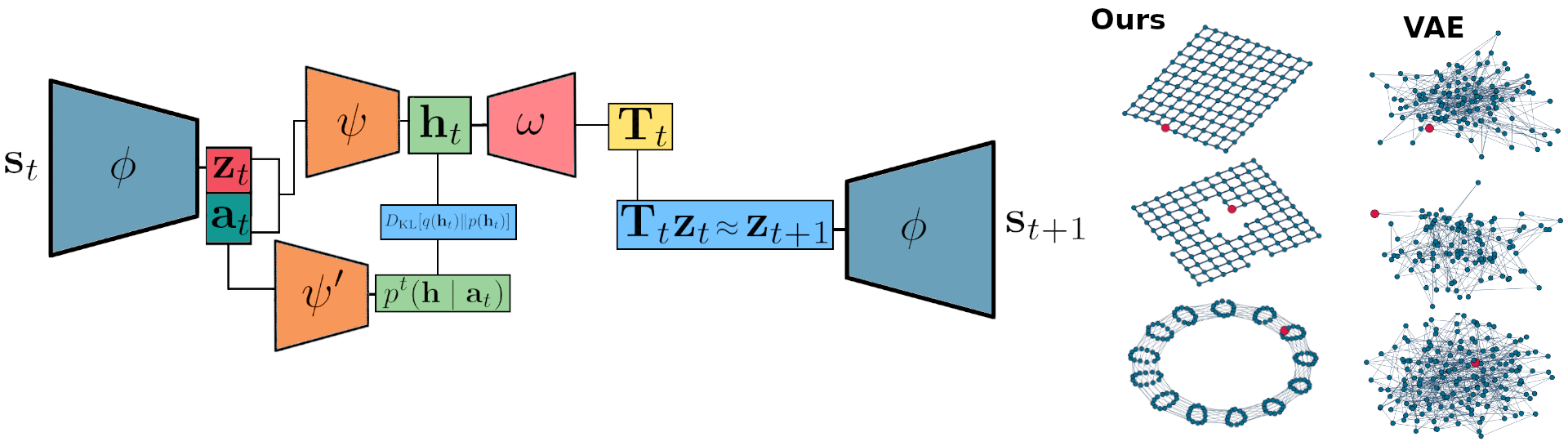}

\end{center}
\caption{Left: Illustration of model components. Right: Embeddings of environment states produced by our model, compared to those produced by a $\beta$-VAE. The VAE embeddings are learnt without parsimony and with a reconstruction objective, making the latent space reflect the topology of the observation space, rather than that induced by the MDP's transition function.}

\end{figure}

Our model learns an encoding function $f_\phi$ that maps states  $\state$ to latent states $\latent$. For the subsequent experiments we assume a deterministic function, but it is straightforward to formulate our model as a stochastic latent dynamics model too (see Appendix \ref{A:transition}). In fact, in section \ref{sec:DML} we perform pixel prediction with a stochastic variant of our model. Given an action $\action$ and the current latent state $\latent$, we seek to predict the next latent state $\inflatentnext$. We represent $\inflatentnext$ as the product of the current latent state with a linear transformation matrix $\latentnext\approx\trans\latent$ which we predict from the latent state-action tuple $(\latent, \action)$.

\begin{equation}
    \begin{aligned}
        &\latent = f_\phi(\state) \\
        &\latenta = g_\psi(\latent, \action) \\
        &\trans = j_\omega(\latenta, \action) \\
        &\inflatentnext = \trans\latent \\
    \end{aligned}    
\end{equation}

We use a probabilistic approach to learning $\trans$: We seek to infer the posterior distribution of a discrete latent code $\latenta \sim q_{\psi}^t(\mathbf{z}_t, \mathbf{a}_t)$, given a prior $p_{\psi}^t(\mathbf{h}\mid \mathbf{a}_t)$ from which we can decode the appropriate transformation $\trans$ that describes the transition. 

\begin{equation}
    \begin{aligned}
        &\text{Posterior: } \ &&q_{\psi}^t(\mathbf{h}\mid \mathbf{z}_t, \mathbf{a}_t) \\
        &\text{Prior: } \ &&p_{\psi'}^t(\mathbf{h}\mid \mathbf{a}_t) \\
    \end{aligned}    
\end{equation}

We want our model to be able to recapitulate observed transitions as accurately as possible, while maximizing the predictability of the transformations describing individual transitions in latent space from the chosen actions alone. This is what we refer to as the principle of parsimony. We construct $q_{\psi}^t(\mathbf{h}\mid \mathbf{z}_t, \mathbf{a}_t)$ as a multivariate Bernoulli distribution of dimensionality $n$ with probability vector $\mathbf{p}_t$, which is predicted by a neural network function $g_{\psi}(\latent, \action)$. The latent code $\latenta$ is a binary vector which we produce by rounding $\mathbf{p}_t$.

\begin{equation}
    h_i = \begin{cases}
    1,& \text{if } p_i \geq 0.5\\
    0,              & \text{otherwise}
\end{cases}
\end{equation}

The latent transition code $\latenta$ is decoded with a learnt decoding function $j_\omega$ into the parameters of the linear transformation matrix $\trans$ with which we predict the next latent state. We then train the transition encoder and decoder networks $g_{\psi}$ and $j_\omega$ using a variational objective \citep{kingma2013auto, hafner2019learning}. To backpropagate through the discrete transition representation, we make us of the straight-through estimator \citep{bengio2013estimating}. Due to the discretization, this transition representation already acts as a bottleneck on the number of transformation matrices we can associate with each action \citep{van2017neural}. However, a chief aim of our model is to represent the transitions of the environment with as \emph{few} transformations as possible. We incorporate this desideratum in the way we construct the variational objective for posterior inference of $q_{\psi}^t(\mathbf{h}\mid \mathbf{z}_t, \mathbf{a}_t)$. We leverage a second learnt neural network function $g_\psi'(\latent, \action)$ to learn a prior distribution $p_{\psi'}^t(\mathbf{h}\mid \mathbf{a}_t)$ where $\latenta$ does not depend on the current latent state $\latent$.

Importantly, doing so makes our prior over transformation matrices, given an action, state-invariant. By enforcing closeness to the prior, we encourage our model to learn a latent state space such that latent state transitions may be predicted accurately even without information about the current latent state $\latent$. Finally, we can construct a loss function for learning the posterior distribution as follows:

\begin{equation}\label{eq:loss_dyna}
    \mathcal{L}_{transition} = \underbrace{\log p(\latentnext\mid \latenta, \action)}_\text{next state prediction} + \underbrace{\beta D_{KL}\left[ q_{\psi}^t(\mathbf{h}\mid \mathbf{z}_t, \mathbf{a}_t) \lVert p_{\psi'}^t(\mathbf{h}\mid \mathbf{a}_t)\right]}_\text{parsimony}
\end{equation}

The first term reflects the accuracy with which our model predicts the next latent state (as encoded by our model) and is a function of the distance between $\inflatentnext$ and $\latentnext$ (see Appendix \ref{A:transition}). The second term reflects how close our posterior over $\latenta$ is to our state-invariant prior, scaled by the hyperparameter $\beta$.

\subsection{Parameterizing transformations}

We consider three types of transformations -- rotations, translations and their composition. As such, we assume that all transitions can be represented as an affine transformation of the current latent state. We leverage a decoder network $j_\omega$ which takes the current latent transition code and action tuple $(\latenta, \action)$ and predicts the parameters of either a rotation matrix $\mathbf{R}$ or a translation matrix. For translations we predict a vector of displacement values $\mathbf{v}$ and predict the next latent state $\inflatentnext = \latent + \mathbf{v}_t$, requiring $n$ parameters for a latent state space of dimensionality $n$. Parameterizing rotations is more involved. \cite{quessard} parametrize rotations with a product of $\frac{n(n - 1)}{2}$ 2-dimensional rotations. In our approach we predict the entries of a skew-symmetric matrix. The space of skew-symmetric matrices form the Lie algebra of the special orthogonal group and its elements can thus be viewed as infinitesimal rotations \citep{sola2018micro}. By taking the matrix exponential of an $n\times n$ skew-symmetric matrix, we obtain an $n\times n$-dimensional rotation matrix $\mathbf{R}$. Since the upper triangle of a skew symmetric matrix is the negative of the lower triangle, we can parameterize the $n$-dimensional rotation using the same number of parameters as \cite{quessard}.

\subsection{Learning the encoding function}

Encouraging the latent dynamics to be parsimonious through the KL term in equation 3 is not sufficient, we also need to make sure that the condition of parsimony is not fulfilled vacuously, for instance if the encoder maps all states $\state$ to a single latent state. A popular approach for avoiding state collapse is to equip the model with a state decoder that tries to predict the state $\state$ from the latent state $\latent$ \citep{hafner2019learning, watter2015embed}. However, encouraging the model to learn latent states that are easily decodable could conflict with our goal of learning a latent state space governed by parsimonious dynamics, as generative factors of the data distribution could influence the topology of our latent space. Instead, we opt for a contrastive objective to distinguish between states \citep{oord2018representation, laskin2020curl}, while giving our transition model the freedom to embed states such that transitions between them can be encoded parsimoniously. Our approach is inspired by noise contrastive estimation \citep{gutmann2010noise, oord2018representation}, which seeks to keep a latent state $\latent\mid\state$ predictable from the observed state, while keeping it diverse from the distinct latent states. For a mini-batch $\mathcal{B}$ we construct target labels for each state $\state\in \mathcal{B}$ and corresponding similarity ratings for our encodings of them:

\begin{equation}
    \begin{gathered}
        l(\mathbf{s}, \mathbf{s}')=e^{-\tau_\mathbf{s}\lVert \mathbf{s} - \mathbf{s}'\rVert_2} \\
        k(\mathbf{z}, \mathbf{z}')=e^{-\tau_\mathbf{z}\lVert  \mathbf{z} - \mathbf{z}'\rVert_2}
    \end{gathered}
\end{equation}

where $l(\mathbf{s}, \mathbf{s}')$ are targets and $k(\mathbf{z}, \mathbf{z}')$ are latent state similarities. Here $\tau_\mathbf{s}$ and $\tau_\mathbf{z}$ are scaling parameters quantifying how quickly state similarity decays with distance. To keep latent states diverse we minimize the cross entropy between $k(\mathbf{z}, \mathbf{z}')$ and $l(\mathbf{s}, \mathbf{s}')$:

\begin{equation}
    \mathcal{L}_{contrastive} = - \dfrac{1}{N}\sum_{\mathbf{s}, \mathbf{s}'\in\mathcal{B}\times\mathcal{B}}k(\mathbf{z}, \mathbf{z}')\log l(\mathbf{s}, \mathbf{s}') + (1 - k(\mathbf{z}, \mathbf{z}'))\log(1-l(\mathbf{s}, \mathbf{s}'))
\end{equation}

By scaling $\tau_\mathbf{s}$ sufficiently high, we encourage our model to distinguish between states that are not identical. This facilitates the learning of parsimonious dynamics: as we only require that distinct states are encoded far enough apart, the transition model is afforded freedom to embed states so that the condition of parsimony is satisfied. Our final loss function is then the sum of the transition loss and the contrastive loss.


\section{Related work}

\textbf{World models}\\
\cite{quessard} represent latent state transitions as the product of the current latent state with elements of the special orthogonal group, i.e. a learnt rotation matrix. However, they assume that the rotations describing state transitions are state-invariant. That is, actions can only affect the state in a single way. We make state-invariance a soft constraint by keeping transition representations close to a state-invariant prior, and represent latent state transitions using elements from the affine group. \cite{watter2015embed} learn to embed high-dimensional inputs in a low-dimensional space in which the dynamics are locally linear, allowing them to plan with stochastic optimal control. Unlike our approach, they use a reconstruction objective to mitigate state collapse, and do not regularize the state-action representations the agent learns. \cite{hafner2020mastering}, \cite{hafner2019dream}, \cite{kaiser2019model} and \cite{ha2018recurrent} learn world models with the purpose of learning policies, either by training the agent within the world model entirely, or by extracting useful latent features of the environment.

\textbf{Planning}\\
Dynamics models are also pervasive in planning tasks. \cite{deisenroth2011pilco} use Gaussian process regression to learn environment dynamics for planning in a sample efficient manner. \cite{schrittwieser2020mastering} learn a latent state dynamics model without a reconstruction objective to play chess, shogi and Go using Monte Carlo Tree Search. \cite{hafner2019learning} learn a recurrent state space model, representing latent states both with a deterministic and stochastic component, and perform planning in pixel environments using the Cross Entropy method. Our approach extends on previous work by building latent state spaces that facilitate planning with incomplete knowledge of the environment. This affordance is due to the latent state space being organized such that transitions can be described with a sparse latent code. 

\textbf{Disentangled representations}\\
Learning disentangled representations is a popular approach for building latent variable models \citep{higgins2016beta, burgess2018understanding}.\cite{higgins2018towards} propose a group theoretic definition of disentangled representations and \cite{caselles2019symmetry} argue that learning symmetry based disentangled representations requires interactions with the environment. Our model can be viewed as learning a latent state space whose dynamics are described by a small number of transformations belonging to the affine group.

\textbf{Navigation}\\
We argue that the ability to navigate could be supported by learning parsimonious dynamics. Using an RL framework, \cite{mirowski2016navigation} learn to navigate to goal locations in 3D mazes with pixel inputs, for which they rely on auxiliary depth prediction and location prediction tasks. \cite{banino2018vector} use velocity and head direction information to learn map-like representations of environments, allowing them to generalize about novel shortcuts to goals. However, biological agents do not usually get information about how close they are to boundaries, their direction of travel, or what their current spatial location is. Rather, these variables must be inferred from interactions with stimuli of much higher dimensionality. We propose a method for learning the latent spatial layout of environments and generalize about their transition dynamics from the principle of parsimony alone, without supervision or additional information about the aforementioned latent variables.

\section{Experiments}

We test our model's ability to learn latent spaces and dynamics that are useful for policy learning and planning. We designed three environments with different topological properties. In all environments the agent is tasked with navigating to a fixed goal location from a fixed starting location, and once it has arrived at the goal location, stay there for the remainder of the episode. The action space consists of the five actions $\mathcal{A} = \lbrace LEFT, RIGHT, UP, DOWN, STAY\rbrace$. The actions are represented as one-hot encoded vectors to not reveal any information about the transition function of the environment. The agent moves around on the grid by selecting a cardinal direction, which moves the agent one unit in the respective, latent direction. The latent coordinate features of the states are unobservable to the agent. With the $STAY$ action the agent stays put on the current state. 

Environment states are represented as random vectors drawn from a multivariate Gaussian $\mathbf{s}\sim\mathcal{N}(\boldsymbol{\mu}, \mathbf{I})$ with a diagonal covariance matrix. These are the vectors that the agent `observes' when occupying a state, and \emph{not}, for instance, a top-down view of the environment. The vectors are drawn when environments are initialized, and then remain fixed for the duration of an experiment. Generating the state vectors from an isotropic Gaussian makes the observation space independent of the underlying hidden variable describing the agent's position. This is a key property of our tasks: We maintain that the ability to learn the group properties of an environment in a way that is disassociated from learning the generative factors of the observations that the states emit is important. Generally, the manifold that the state-observations lie on may be entirely different from the manifold defined by an environment's transition function. We hypothesize that structuring the latent state space to reflect the topology of the environment is beneficial for solving several RL tasks.

\subsection{Gridworlds}

We designed two $11\times11$ state discrete gridworlds (see Figure \ref{fig:mfrl}). On the boundary states of the gridworld there were walls. One of the gridworlds was partitioned into four rooms by walls. Information about whether the agent was facing a wall in any of the four cardinal directions was encoded as a binary vector, which we concatenated with the initial random state vector to produce what the agent sees when occupying a state. 

\subsection{Torus}

\begin{figure}[t!]
\begin{center}
\includegraphics[width=1\textwidth]{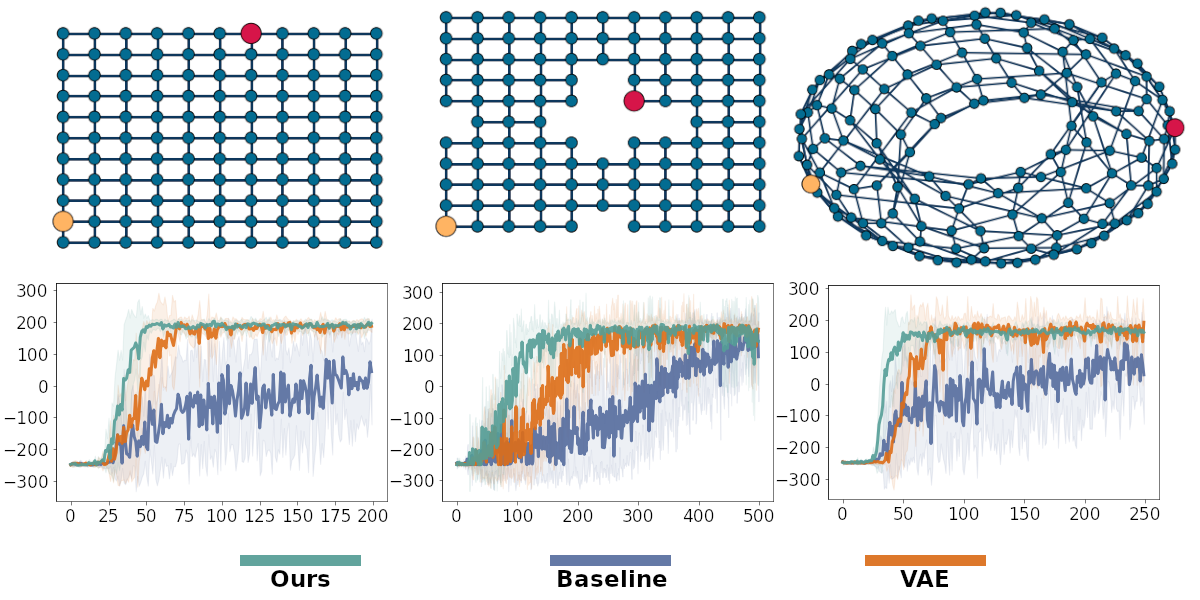}

\end{center}
\caption{Top: The environments have a starting state (yellow) and a goal state (red). Bottom: The corresponding learning curves of a SAC agent learning policies over three types of latent state spaces, averaged over 10 seeds: \textit{Ours} has a latent state space constrained to be such that dynamics are parsimonious. \textit{Baseline} has no representation learning objective. \textit{VAE} learns a latent state space using the $\beta$-VAE model, with next state prediction as an auxiliary task.}
\label{fig:mfrl}
\end{figure}

We designed a discrete torus world similar to the gridworld by connecting the gridworlds boundary states to the corresponding boundary states on the opposite end (see Figure \ref{fig:mfrl}). The torus contained $13\times 13$ states and no boundaries.

\subsection{Model free learning}

We designed a model free learning task for each environment. In each task, the agent needs to learn to move from a starting state $\mathbf{s}_{start}$ to an unknown goal state $\mathbf{s}_{goal}$ and stay there for the remainder of an episode which lasted for 250 timesteps. Each state except for the goal state yields a reward of $-1$ when exited except for the goal state which yields a reward of $+1$. The agent learns a policy which takes it to the goal location using the Soft Actor Critic (SAC) algorithm \citep{haarnoja2018soft} adapted for discrete action spaces \citep{christodoulou2019soft} (see Appendix \ref{A:sac} for details). Crucially, the agent learns a policy over its latent state representations $\pi(\mathbf{a}\mid\mathbf{z})$ as opposed to the observations the environment emits. We train agents for 200 episodes in the gridworld, 500 episodes in the four rooms environment, and for 250 episodes in the torus world. Each episode lasts for 250 steps. The agents are trained as described in Algorithm \ref{algo:mfrl}. We compare our model to two alternative representation learning approaches: i) The $\beta$-Variational Autoencoder \citep{higgins2016beta, lee2020stochastic} which learns disentangled probabilistic embeddings by reconstructing the states as well as performing next state prediction. Next state prediction is performed by multiplying the current latent state with a predicted affine transformation matrix, however, without regularizing the state-action representation to be parsimonious. ii) A baseline feedforward neural network for which no representation learning objectives influenced latent state representations except the actor and critic losses. Our model achieves the best total score summed over episodes in all environments, averaged over 10 seeds. We fine tuned the regularization coefficient $\beta$ in equation \ref{eq:loss_dyna} and the $\beta$ of the VAE model. Comparisons revealed that regularizing the latent state-action representation $\latenta$ proved beneficial in all environments except the Four Rooms environment in which a $\beta$ value of 0 proved slightly better than the next best regularized implementation of our model.  

\subsection{Planning}

For each environment we generated a set of planning problems where the agent starts in a random state, and needs to plan a sequence of actions to reach a goal state and stay there for the remainder of an episode which lasted for 50 timesteps. The agent has no knowledge of environment dynamics initially except for what the goal state is, and needs to learn a viable dynamics model as it engages with the task. The agent attempts to solve the planning problems by encoding the goal state into its learnt latent state space $\mathbf{z}_\text{goal}$ and by simulating trajectories that take it to the goal. The agent estimates the return of a trajectory as the sum of latent state occupancies weighted by the  exponential of their negative distance to the latent goal state $G = \sum_{t=0}^H e^{-\lVert\latent - \mathbf{z}_{goal}\rVert_2}$. After an episode, the agent fits its dynamics model to the observations it gathered through executing its plan. Each planning task consists of 30 such planning problems, varying with difficulty, as some goal locations are further away from the agents' starting location. Following \cite{hafner2019learning}, we use the Cross Entropy Method (CEM) \citep{rubinstein1999cross} as our planning algorithm (see \ref{algo:CEM}). We verified that it was able to solve most tasks when using the true environment dynamics with a moderate planning budget.

We compare our model to alternative latent dynamics models: A deterministic RNN and a stochastic latent state model \citep{hafner2019learning} trained with a $\beta$-VAE objective \citep{higgins2016beta}. All models represent state transitions as the product of the current latent state with a learnt affine transformation matrix, but lack the parsimony constraint at the core of our model. The models were trained as described by Algorithm \ref{algo:planning}. As with the policy learning task, we found that our model achieved the best score pooled over the 30 planning tasks and 10 seeds. Moreover, regularizing $\latenta$ proved beneficial in all environments, providing further evidence that parsimonious dynamics are beneficial for planning. 

\begin{figure}[t!]
\begin{center}
\includegraphics[width=1\textwidth]{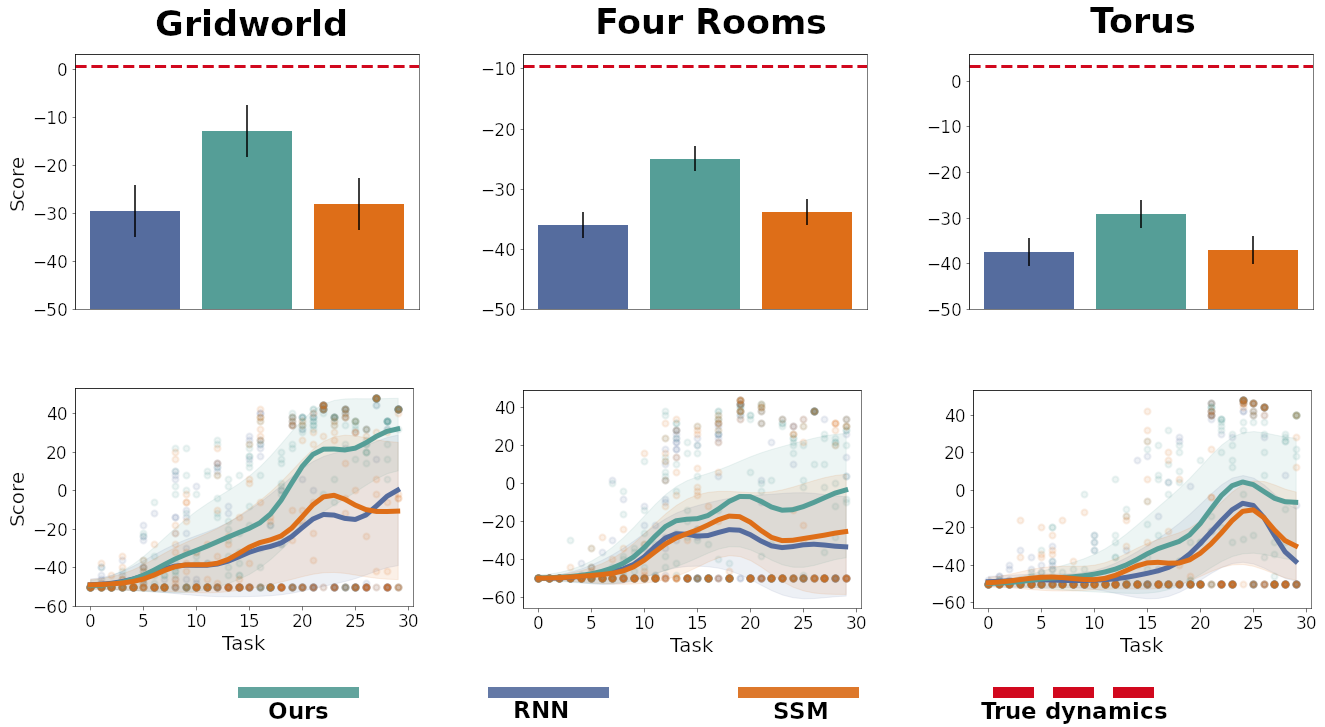}

\end{center}
\caption{Top row: The score achieved by the planning agent using different dynamics models, averaged over the 30 tasks and 10 seeds: Our approach beats the deterministic recurrent world model (RNN) and the stochastic world model (SMM) trained with a disentanglement objective. Error bars show standard deviation computed across seeds. Red lines show score of planning agent using the true environment dynamics. Bottom row: The score achieved for each task, dots indicate individual samples across 10 seeds, and curves are smoothed using a Gaussian filter with standard error computed across seed, with lengthscale $\sigma = 2$.}

\end{figure}

\section{Learning parsimonious dynamics from pixels}\label{sec:DML}

We sought to evaluate our model's ability to perform long-term future state prediction in an environment with pixel inputs. For this task, we relied on the Deepmind Lab environment, a challenging partially observable environment with image observations \citep{beattie2016deepmind}. To make our model suitable for pixel prediction, to mitigate the partially observability, and to make it comparable to other models in the literature, we used the stochastic variant of our model (see Appendix \ref{A:transition}) and image reconstruction loss rather than a contrastive objective to avoid latent state collapse. Furthermore, we endowed it with a convolutional neural network image encoder, a transpose convolutional neural network decoder, and recurrent neural network whose outputs were concatenated with the inferred latent state for pixel prediction (see Appendix \ref{A:dml}).

As a comparison model we chose the Recurrent State Space Model (RSSM) from \cite{hafner2019learning} (see Appendix \ref{A:rssm}). When applicable, we also used the hyperparameters they provide for our model. We trained both models to reconstruct sequences of images, conditioned on previous image and action observations, collected from an agent executing a random policy for 250 episodes in the \texttt{seekavoid\_arena\_01} environment. No velocity or location information was provided to the agent. We then made the models perform open-loop prediction of 30 test sequences of 149 environment steps, that were not in the training set. We evaluated open-loop reconstruction errors and the KL divergence between predicted future latent states and closed-loop inferred latent states from observations $KL_D\left[q(\inflatentnext\mid \textbf{z}_{\leq t}, \textbf{u}_{\leq t}, \textbf{a}_{\leq t})\rVert p(\latentnext\mid \textbf{s}_{t+1}\textbf{u}_{t})\right]$, where $\mathbf{u}_t$ is a deterministic latent variable provided by the recurrent neural network. Our model generalized better to the test sequences, achieving both lower KL divergences between predicted and encoded states, and lower average reconstruction error in the open-loop prediction task. This provides evidence for the utility of learning parsimonious dynamics in more challenging pixel environments as well.

\begin{figure}[t!]
\begin{center}
\includegraphics[width=1\textwidth]{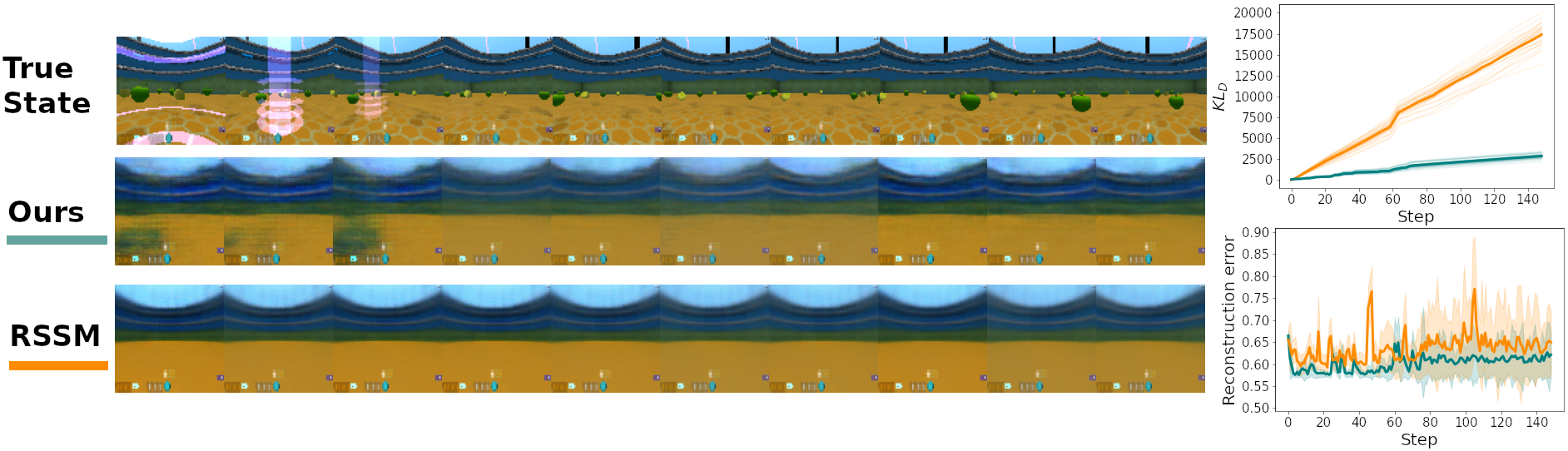}

\end{center}
\caption{Open loop prediction of future states in the Deepmind Lab environment. Our model is better at predicting transitions in latent space, and reconstructing images from imagined latent states.}

\end{figure}

\section{Discussion}

In the current paper, we introduced a model that learns a parsimonious set of transformation matrices to describe the dynamics of MDPs in a latent state space. Learning a world model whose states are organized according to this principle proved beneficial in a policy learning task, a planning task and a video prediction task. With the objective of carving the environment at its joints rather than the observations its states emit, the learnt latent states contained information that was valuable for policy learning. Moreover, planning in the learnt latent space became feasible having observed fewer state transitions: This is because the agent could systematically generalize about the dynamics of parts of the environment that were not yet explored exhaustively. We have shown that simply endowing the dynamics model's objective with a term encouraging parsimony was sufficient to produce latent spaces that display useful characteristics.

We investigated the utility of parsimonious dynamics in simpler environments whose transitions could indeed be characterized by a small set of linear transformations. A limitation of our approach is that the environments we investigated were rather simplistic compared to the rich environments humans and other state-of-the-art models have been shown to be able to navigate through. To remedy this we provided promising initial evidence that parsimonious dynamics can facilitate future state prediction in a Deepmind Lab environment. In future work we intend to scale up our approach to also perform policy search in similarly rich environments. A further limitation is that the degree of sparseness with which the model tries to recapitulate transitions is controlled by the KL scaling term $\beta$. Though we avoid assuming that dynamics are completely state-invariant, we still have to tune $\beta$. In future work, we seek to address this issue by adaptively regulating the complexity of our learnt dynamics to reflect the complexity of the environment.

The principle of parsimony was initially motivated from the viewpoint of cognitive science. \cite{tolman1948cognitive} showed that rats preferred a novel shortcut over a repeatedly reinforced longer route to a goal location, hypothesizing that animals learn and use latent representations of their environment that must contain assumptions about how they are structured. Seminal work in neuroscience demonstrated the existence of neurons selectively tuned to specific spatial positions (place cells), and others that represent global geometrical information about the environment (grid cells), representations that were also found in artificial agents trained to navigate in Euclidean spaces \citep{banino2018vector, cueva2018emergence}. Recent cognitive neuroscience studies revealed that humans rely on cognitive maps to navigate complex environments \citep{epstein2017cognitive}, abstract spaces \citep{constantinescu2016organizing, garvert2017map}, generalize about rewards \citep{garvert2021hippocampal}, and draw inferences about transition dynamics in novel environments \citep{mark2020transferring}. Future work could also investigate the role of parsimony in the mental maps that humans and other animals build of the worlds they inhabit.

\newpage

\bibliography{iclr2023_conference}
\bibliographystyle{iclr2023_conference}

\newpage
\appendix
\section{Appendix}

\subsection{Transition loss}\label{A:transition}

Our model can be formulated as both a deterministic and stochastic state space model. The loss function for deterministic transitions is described in Equation \ref{eq:loss_dyna}. In the stochastic case, we produce probabilistic state embeddings $\latent \sim p(\latent\mid \state)$, where $p$ is a Gaussian with mean and standard deviation $(\boldsymbol{\mu}_t, \boldsymbol{\sigma}_t) =f_\phi(\state)$. We construct a lower-bound on the data-log-likelihood by minimizing either a reconstruction error term or a contrastive term, in addition to a KL term constraining the posterior latent state to be close to the prior. 

\begin{equation}
    \mathcal{L} = \log p(\state\mid\latent) +  \beta D_{KL}\left[q(\latent\mid \state) \lVert p(\tilde{\mathbf{z}}_t\mid \mathbf{z}_{t-1}, \mathbf{a}_{t-1})\right]
\end{equation}


We allow gradients to pass through the stochastic component using the reparameterization trick \citep{kingma2013auto}.
\subsubsection{Deterministic transitions}

If our transition model is deterministic, we define the reconstruction term based on the exponential of the norm of the vector $\mathbf{z}_{error} = \inflatentnext - \latentnext$:

\begin{equation}
    \mathcal{L}_{\text{deterministic transitions}} = \Vert\inflatentnext - \latentnext\rVert_2^2 + e^{-\Vert\inflatentnext - \latentnext\rVert_2}
\end{equation}

We found that adding the mean squared error term in the transition reconstruction helped the models converge.
\subsubsection{Stochastic transitions}

In the stochastic case, we use as our transition reconstruction term the KL divergence between the two distributions predicted for the next latent state: $\mathcal{N}(\mathbf{T}_t\boldsymbol{\mu}_t, \tilde{\boldsymbol{\sigma}}_{t+1})$ where $\tilde{\boldsymbol{\sigma}}_{t+1}=g_\psi(\boldsymbol{\mu}_t, \action)$ and $\mathcal{N}(\boldsymbol{\mu}_{t+1}, \boldsymbol{\sigma}_{t+1}) = f_\phi(\mathbf{s}_{t+1})$, e.g.:

\begin{equation}\label{eq:stochastic}
    \mathcal{L}_{\text{stochastic transitions}} = D_{KL}\left[p(\latentnext) \lVert q(\inflatentnext)\right] + D_{KL}\left[ q(\mathbf{h}\mid\mathbf{z}_t, \mathbf{a}_t) \rVert p(\mathbf{h}\mid \mathbf{a}_t)\right]
\end{equation}
Our composite loss function therefore becomes the standard variational objective for training the stochastic state space model, plus the KL term describing the divergence between our prior state action representation $p(\latenta)$ and our posterior $q(\latenta)$. 

\subsection{Dynamics models}\label{A:dynamics}
For the policy learning and planning experiments we parameterized our dynamics models using feedforward neural network encoders and decoders (when applicable), each with two hidden layers and 1200 ReLU units \cite{nair2010rectified}. The latent space the encoders projected to had 15 dimensions. The state-action variable $\latenta$, which also had 15 dimensions, was learnt using an encoder and decoder with the same hyperparameters. When we used the contrastive loss to avoid state collapse, we used similarity scalings $\tau_{\mathbf{s}} = 100$ and $\tau_{\mathbf{z}} = 0.1$


\subsection{Deepmind Lab}\label{A:dml}

We trained our dynamics models on $80\times80$ resolution image sequences from Deepmind Lab's \texttt{seekavoid\_arena\_01}. The sequences were produced by an agent with a random policy. The agent could execute six discrete actions: Move forward, backwards, left or right, and turn in both horizontal directions. We equipped our model with a convolutional neural network for image encodings and a transposed convolutional neural network for image reconstructions, both with the architecture from \cite{ha2018recurrent}. For our recurrent neural network producing the deterministic latent variable $\mathbf{u}_t$ we used a GRU \citep{chung2014empirical} with 200 hidden units. The feedforward neural networks which were used as encoders and decoders had all two hidden layers with 200 ReLU units.

Like in \cite{hafner2019learning}, we train our model on sequences of image-action pairs: To perform one gradient step with our dynamics models we draw $B=50$ sequence chunks of length $N=50$ and compute the dynamics loss using Equation \ref{eq:stochasticloss}:

\begin{equation} \label{eq:stochasticloss}
\begin{split}
    \mathcal{L} &= \underbrace{\log p(\state\mid\latent, \mathbf{u}_t)}_{\text{reconstruction loss}} \\
    &+  \beta_1 \underbrace{D_{KL}\left[q(\inflatentnext\mid \latent, \mathbf{u}_t, \action) \lVert p(\latentnext\mid \mathbf{s}_{t+1}, \mathbf{u}_t)\right]}_{\text{transition loss}} \\
    &+ \beta_2 \underbrace{D_{KL}\left[ q(\mathbf{h}\mid\mathbf{z}_t, \mathbf{u}_t,\mathbf{a}_t) \rVert p(\mathbf{h}\mid \mathbf{u}_t, \mathbf{a}_t )\right]}_{\text{parsimony}}
\end{split}
\end{equation}

The latent variables $\latent$ and $\mathbf{u}_t$ had 30 dimensions each. We perform 25 gradient steps per episode, with the transition KL term $\beta_1=0.1$ and the parsimony KL term $\beta_2=0.5$.

\subsection{RSSM}\label{A:rssm}

We implemented the RSSM from \cite{hafner2019learning}. The RSSM consists of the following components:
\begin{equation}
    \begin{aligned}
    &\text{Deterministic state model} \ &&\mathbf{h}_{t+1} = f(\latent, \mathbf{h}_t, \action) \\ 
    &\text{Stochastic state model} \ &&\mathbf{z}_{t+1} \sim p(\mathbf{h}_t) \\ 
    &\text{Observation model} \ &&\state \sim p(\latent, \mathbf{h}_t) \\ 
    \end{aligned}
\end{equation}

where $ f(\latent, \mathbf{h}_t, \action)$ is a GRU with 200 hidden dims. The RSSM is trained similarly to our model above, performing gradient steps on reconstructions of batches of image sequences. The loss function used to train the RSSM is the following:

\begin{equation} \label{eq:rssmloss}
\begin{split}
    \mathcal{L} &= \underbrace{\log p(\state\mid\latent, \mathbf{h}_t)}_{\text{reconstruction loss}} \\
    &+  \beta_1 \underbrace{D_{KL}\left[q(\latent\mid \state, \mathbf{h}_t) \lVert p(\latent\mid \mathbf{h}_t)\right]}_{\text{transition loss}} 
\end{split}
\end{equation}

Identically to our model, the latent variables $\latent$ and $\mathbf{h}_t$ had 30 dimensions each. Image encoding and decoding was done with the same convolutional neural network architectures described in the previous section. We perform 25 gradient steps per episode, drawing $B=50$ sequence chunks of length $N=50$, with the transition KL term $\beta_1=0.1$.

\subsection{Policy learning}

Agents were trained to maximize future reward using the actor-critic method. Both actors and critics were parameterized by feedforward neural networks with an encoder and decoder, each of which consisted of two hidden layers with 800 ReLU units. The policy was learnt using the following algorithm:  

\begin{algorithm}[H]\label{algo:mfrl}
\caption{Policy learning}
\begin{algorithmic}
\For{each episode}
\For{each environment step}
    \State $\latent = f_{\phi}(\state)$
    \State $\action\sim \pi_{\theta_{\text{actor}}}(\latent)$
    \State $r_t, \mathbf{s}_{t+1} \sim $ \texttt{Env.step}$(\state, \action)$
    \State $\mathcal{D} \gets \mathcal{D}\ \cup (\state, \action, r_t, \mathbf{s}_{t+1})$
\EndFor
\For{each dynamics gradient step}
    \State Sample batch of $N$ tuples $(\state, \action, \mathbf{s}_{t+1})\sim \mathcal{D}$
    \State Compute loss $\mathcal{L}_{\text{dynamics}}$ \Comment{Use objective of the relevant dynamics model}
    \State $\theta_{\text{dynamics}} \gets \theta - \alpha\nabla \mathcal{L}_{\text{dynamics}}$
 
\EndFor
\For{each policy gradient step}
    \State Sample batch of $M$ tuples $(\state, \action, r_t, \mathbf{s}_{t+1})\sim \mathcal{D}$
    \State Compute actor and critic losses $\mathcal{L}_{\text{actor}}, \mathcal{L}_{\text{critic}}$ \Comment{Use soft actor critic losses}
    \State $\theta_{\text{actor}} \gets \theta_{\text{actor}} - \alpha\nabla \mathcal{L}_{\text{actor}}$
    \State $\theta_{\text{critic}} \gets \theta_{\text{critic}} - \alpha\nabla \mathcal{L}_{\text{critic}}$
\EndFor

\EndFor
\end{algorithmic}
\end{algorithm}

\subsection{Soft Actor Critic}\label{A:sac}

The actor and critic networks were trained using the Soft Actor Critic algorithm (SAC) \citep{haarnoja2018soft}. To learn Q-values, we used two source critic networks and two target critic networks, fine-tuned the inverse reward scale parameter $\alpha = 0.5$, the number of policy gradient steps $M = 15$, batch size (150 in the gridworld and torus environments, and 350 in the Four Rooms environment) and the target value smoothing update constant $\tau_{target} = 0.1$. We trained the networks with the Adam optimizer \citep{kingma2014adam}, using a learning rate of $1e-4$. These hyperparameters were fine-tuned to maximize performance of the baseline agent (which had no latent dynamics model). We used these hyperparameter settings with other models and did no further fine-tuning, except for optimizing hyperparameters specific to the dynamics models. We trained the dynamics models using a learning rate of $1e-3$, also using the Adam optimizer.

\subsection{Planning}

In the planning task we leveraged the Cross Entropy Method \citep{rubinstein1999cross} in tandem with learnt dynamics models, using hyperparameters similar to \cite{hafner2019learning}. The agents planned over a horizon length $H=15$, with 10 iterations $I$, 1000 samples $J$ per iteration, and updating plans using the $K=200$ best samples. We trained the dynamics models using a learning rate of $1e-3$ with the Adam optimizer. To make sure the agents explored sufficiently, we used the epsilon-greedy heuristic, scheduling $\varepsilon$ so that the agents explored a lot early in the experiment and little late. We made $\varepsilon$ a function of the number of tasks played $\mathcal{E}(N) = 1 - \left(\dfrac{N-1}{T}\right)^V$ where $N$ is the current task number, $T$ the total number of tasks and $V=2.8$ a scaling term. 

\begin{algorithm}\label{algo:planning}
\caption{Planning}
\begin{algorithmic}
\For{each task}
\State $\mathbf{s}_{\text{start}}, \mathbf{s}_{\text{goal}} \sim \mathcal{S}$ \Comment{Draw start and goal states uniformly, s.t. $\mathbf{s}_{\text{start}}\neq \mathbf{s}_{\text{goal}}$}
\State $\mathbf{z}_{\text{goal}} = f_{\phi}(\mathbf{s}_{\text{goal}})$
\State $\tilde{r}(\mathbf{z}) = e^{-\lVert f_{\phi}(\mathbf{s}) - \mathbf{z}_{\text{goal}}\rVert_2}$ \Comment{Define reward function for model}
\State $\varepsilon_i \gets \mathcal{E}(task)$
\For{each environment step}
    \State $\latent = f_{\phi}(\state)$
    \State $\action = \begin{cases}
    \text{CEM}(\latent)& \text{with probability } 1 - \varepsilon_i\\
    \text{Random}              & \text{with probability } \varepsilon_i
\end{cases}$ \Comment{Use Cross Entropy Method}
    \State $r_t, \mathbf{s}_{t+1} \sim $ \texttt{Env.step}$(\state, \action)$
    \State $\mathcal{D} \gets \mathcal{D}\ \cup (\state, \action, \mathbf{s}_{t+1})$
\EndFor
\For{each dynamics gradient step}
    \State Sample batch of $N$ tuples $(\state, \action, \mathbf{s}_{t+1})\sim \mathcal{D}$
    \State Compute loss: $\mathcal{L}_{\text{dynamics}}$\Comment{Use objective of the relevant dynamics model}
    \State $\phi_{\text{dynamics}} \gets \phi - \alpha\nabla \mathcal{L}_{\text{dynamics}}$
 
\EndFor

\EndFor
\end{algorithmic}
\end{algorithm}

Below is a description of the Cross Entropy Method, drawing inspiration from \cite{hafner2019learning}. We adapt it for a discrete action setting, by optimizing the logits of a categorical distribution from which actions are sampled.

\begin{algorithm}\label{algo:CEM}
\caption{Cross Entropy Method}
\begin{algorithmic}
\Require Dynamics model $p(\latentnext\mid\action, \latent)$, reward function $\tilde{r}(\mathbf{z})$
\State Initialize $J$ logits for categorical action distributions $\mathbf{w}\sim \mathcal{N}(0, \mathbf{I})$, where $\mathbf{w}\in \mathbb{R}^{\mid\mathcal{A}\mid\times H}$
\For{each iteration $i=1, ..., I$}
\For{each action sequence $j=1, ..., J$}
    \State $\mathbf{a}_{1:H}\sim\text{Softmax}(\mathbf{w})$ \Comment{Sample actions} 
    \State $\mathbf{z}_{1:H} \sim  \mathbf{p}(\mathbf{z}_{1:H}\mid\mathbf{a}_{1:H-1}, \mathbf{z}_{1:H-1})$ \Comment{Simulate future states with
    dynamics model}
    \State $\mathbf{r} \gets \tilde{r}(\mathbf{z})$ \Comment{Compute rewards using reward function}
\EndFor
\State $\mathcal{K} = argsort\lbrace\mathbf{r}_{j=1}^{J}\rbrace_{1:H}$ \Comment{Get logits of $K$ best performing samples}
\State $\boldsymbol{\mu}_{1:H} \gets \dfrac{1}{K} \sum_{k\in\mathcal{K}} w^{k}_{1:H}$
\State $\mathbf{w}\sim \mathcal{N}(\boldsymbol{\mu}, \mathbf{I})$ \Comment{Sample new logits based on average logits of $K$ best samples}


\EndFor
\end{algorithmic}
\end{algorithm}

\end{document}